\pdfoutput=1

\documentclass[11pt]{article}

\usepackage{hyperref}

\usepackage{EMNLP2022}

\usepackage{times}
\usepackage{latexsym}

\usepackage[T1]{fontenc}

\usepackage[utf8]{inputenc}

\usepackage{microtype}

\usepackage{inconsolata}

\usepackage{graphicx}
\usepackage{subcaption}
\usepackage{comment}
\usepackage{lipsum}

\usepackage{newpxtext}

\makeatletter
\newcommand{\ie}{\emph{i.e.}\@ifnextchar.{\!\@gobble}{}}

\title{
Questioning the Validity of Summarization Datasets and Improving Their Factual Consistency
}

\author{Yanzhu Guo$^1$,  Chloé Clavel$^2$, Moussa Kamal Eddine$^1$, Michalis Vazirgiannis$^1$ \\
$^1$LIX, École Polytechnique, Institut Polytechnique de Paris, France \\ 
$^2$LTCI, Télécom-Paris, Institut Polytechnique de Paris, France \\
\texttt{\{yanzhu.guo, moussa.kamal-eddine\}@polytechnique.edu}\\
\texttt{chloe.clavel@telecom-paris.fr}, \texttt{mvazirg@lix.polytechnique.fr} }

\begin{document}
\maketitle
\begin{abstract}
The topic of summarization evaluation has recently attracted a surge of attention due to the rapid development of abstractive summarization systems. However, the formulation of the task is rather ambiguous, neither the linguistic nor the natural language processing community has succeeded in giving a mutually agreed-upon definition. Due to this lack of well-defined formulation, a large number of popular abstractive summarization datasets are constructed in a manner that neither guarantees validity nor meets one of the most essential criteria of summarization: factual consistency. In this paper, we address this issue by combining state-of-the-art factual consistency models to identify the problematic instances present in popular summarization datasets. We release SummFC, a filtered summarization dataset with improved factual consistency, and demonstrate that models trained on this dataset achieve improved performance in nearly all quality aspects. We argue that our dataset should become a valid benchmark for developing and evaluating summarization systems.
\end{abstract}

\section{Introduction}
While the revolutionary success of the Transformer \cite{NIPS2017_3f5ee243} architecture has drawn a surge of attention to automatic summarization, most research has been focused on improving performance metrics of summarization models on a set of popular datasets. It is often taken for granted that these datasets provide representative examples of high quality summaries, and that models capable of producing summaries that are similar to the ones of the dataset are superior in the task of automatic summarization. \textit{But is this really the case?} 

Numerous works \cite{fabbri-etal-2021-summeval, huang-etal-2020-achieved} have already pointed out deficiencies in the most widely employed summarization datasets (e.g. CNN/DailyMail \cite{NIPS2015_afdec700, nallapati-etal-2016-abstractive} and XSUM \cite{narayan-etal-2018-dont}). These datasets are typically composed of news articles automatically extracted from news websites, paired together with a highlight or introduction sentence which serves as the summary. However, the qualities we seek in highlights and introductions are fundamentally different from the ones we seek in summaries. It is thus time for the NLP community to take a step back, revisit the formulation of the summarization task and reconsider the appropriateness of currently employed datasets.

The goal of automatic summarization is to take an information source, extract content from it and present the most important content to the user in a concise form and in a manner sensitive to the user's or application's needs \cite{mani2001automatic}. In the scope of this paper, we solely focus on generic summarization \cite{nenkova2011automatic}, where we make few assumptions about the audience or the goal for generating the summary. While the importance of the selected content and the conciseness of its form are rather subjective, there is one criterion for summarization that is certain: the summary content should be extracted from the information source. In other words, the summary should be factually consistent with the source document. This is however not the case with summaries constructed from highlights or introductions. It is common for highlights and introductions to contain information that are not mentioned in the main article or even exaggerate certain facts to achieve the goal of click baiting. Therefore, the models trained on these datasets perform the task of ``pitch generation'' rather than the intended summary generation. An example of such a reference summary is given in Table~\ref{table:examples}.

In this paper, we aim to identify these erroneous data samples by scoring them with factual consistency models. We push towards answering two main \textbf{research questions} :

\paragraph{Q1} \textit{What kind of factual inconsistency can different factuality models capture and how can they be combined for better detection of erroneous samples?} This question is answered in Sections \ref{Analysis Using the FRANK Benchmark} and \ref{Filtration} .

\paragraph{Q2} \textit{Does having more reliable datasets with more factually consistent reference summaries lead to better performing summarization models? And do the results only improve in factual consistency or also in other quality aspects such as informativeness and saliency?} We address this question in Section \ref{Experiments and Results}.

Our research aims to answer these questions by filtering out samples with problematic reference summaries. We achieve this by leveraging state-of-the-art factual consistency models. We test the performance of different models on different categories of factuality errors \cite{pagnoni-etal-2021-understanding} and eventually combine them for an optimized filtration methodology. 

Our \textbf{contributions} are summarized below:

\begin{enumerate}

\item We prove the effectiveness of three state-of-the-art factual consistency models in detecting factuality errors. We use these models to affirm the factual consistency issue present in three of the most popular summmarization datasets. We devise a filtration methodology that combines different state-of-the-art factuality models and achieves better detection of misleading reference summary samples. 

\item We release SummFC\footnote{The dataset is publicly available at \url{https://github.com/YanzhuGuo/SummFC}.}, a \textbf{Summ}arization dataset with improved \textbf{F}actual \textbf{C}onsistency. We prove that fine-tuning summarization models on this dataset leads to better performance on not only factuality but also other quality aspects. We believe that regularly amending issues in widely employed benchmark datasets should become a common practice in NLP.

\end{enumerate}

\begin{table}
\begin{center}
\begin{tabular}{ |p{0.94\linewidth}| } 
 \hline
 \textsl{\textcolor{blue}{Source Document: }} \\
 \textsl{Archibald, 22, dominated the event, with Dutch 
rider Kirsten Wild second and Belgium's Lotte 
Kopecky third. That came 24 hours after she won 
her third consecutive women's individual pursuit 
title, having gained silver in Thursday's 
elimination race. The Scottish cyclist won gold in 
the pursuit quartet at the Olympics in Rio.}  \\
 \hline
 \textsl{\textcolor{blue}{Factually Inconsistent Reference Summary: }} \\
 \textsl{british olympic champion  \textcolor{red}{katie} archibald won 
omnium gold at the \textcolor{red}{european track} 
\textcolor{red}{championships}, her second title in two nights 
\textcolor{red}{in paris}.}\\
 \hline
\end{tabular}
 \caption{\label{table:examples}
Example of a reference summary from the XSUM dataset. The words marked out in \textcolor{red}{red} represent factually inconsistent information. The most common type of error for the XSUM dataset is content verifiability errors.}
\end{center}
\end{table}

\section{Related Work}
\paragraph{Summarization Benchmark Datasets}
Summarization benchmark datasets are typically composed of a large number of news documents paired together with ``gold-standard'' human reference summaries. Unfortunately, the human reference summaries in these datasets are often constructed in a suboptimal way. \citet{gehrmann2022repairing} analyze a sample of 20 papers proposing summarization approaches published in 2021; they find 27 datasets that models were being evaluated on. The most popular ones, CNN/DM \cite{nallapati-etal-2016-abstractive} and XSum \cite{narayan-etal-2018-dont}, were used five and four times respectively. However, both of these datasets have multiple issues. Taking the CNN/DM dataset for example, the construction is done by pairing an article with the bullet points written for it on the CNN and Daily Mail websites. This design works well for its initial use as a Question Answering dataset \cite{NIPS2015_afdec700}, but does not function at the same level after its adaptation for summarization. Regarding the XSUM dataset, the main issue concerns its factuality. The reference summaries were never meant to be a real summary, there is thus no requirement for it to be faithful to the source article. An analysis of XSum finds that over 70\% of reference summaries contain factual inconsistencies \cite{maynez-etal-2020-faithfulness}. 

\paragraph{Factual Consistency Models}
Thanks to the development of reference-free factual consistency metrics for summarization \cite{huang2021factual}, the evaluation of factual consistency can now be performed automatically on a large scale. Since such metrics are generally model-based, we refer to them as factual consistency models, in order to distinguish them from the evaluation metrics in Section \ref{Evaluation Metrics}. Popular factual consistency models can generally be categorized into two different paradigms: Question Answering \cite{durmus-etal-2020-feqa, wang-etal-2020-asking} and Entailment Classification \cite{kryscinski-etal-2020-evaluating, goyal-durrett-2020-evaluating}. Question Answering models are based on the intuition that if we ask the same question to a summary and its source article, they should provide similar answers if the summary is factual. Entailment classification models rely on the idea that a factually consistent summary should be semantically entailed by the source article. They are usually trained models fine-tuned on either synthetic or human-annotated datasets. According to the FRANK benchmark \cite{pagnoni-etal-2021-understanding}, Entailment classification models perform significantly better than the Question Answering ones. In addition, some text generation evaluation metrics \cite{zhang2019bertscore, NEURIPS2021_e4d2b6e6} falling out of these two paradigms are also proven to perform well as factual consistency models \cite{pagnoni-etal-2021-understanding}.

\paragraph{Improving Factual Consistency} The idea defended in this paper is that factual consistency models allow us to discover issues in current datasets and eventually release improved versions. Dataset quality in summarization has only started receiving attention recently and only a few papers have attempted to make efforts in this direction. \citet{gehrmann-etal-2021-gem} release an improved version of XSUM that filters the dataset with a BERT-based classifier fine-tuned on 500 document-summary pairs, manually annotated for faithfulness \cite{maynez-etal-2020-faithfulness}. However, the classifier is fairly naive with only a single classification layer added after the BERT model and they did not compare the performance of models trained respectively on the original and improved datasets. Along similar lines, \citet{matsumaru-etal-2020-improving} build a binary classifier for detecting untruthful article-headline pairs and filter a headline generation dataset. \citet{nan-etal-2021-entity} also perform filtration for summarization datasets but both the filtration and evaluation methodologies are limited to the entity level. \citet{goyal-durrett-2021-annotating} identify non factual tokens in the XSUM training data with the Dependency Arc Entailment (DAE) model. They mask these tokens corresponding to unsupported facts and ignore them during the training step. While this approach does allow for improved summarization quality, we see potential issues arising when isolated ignored tokens are confronted with the contextual nature of the Transformer architecture. Filtering out entire samples with detected inconsistency is thus the preferred solution. 

We present the first work on a systematic investigation of an optimal methodology for improving factual consistency in summarization datasets. We are also the first to comprehensively analyze the effect of training data quality on summarization system outputs. We furthermore extend the range of analysis compared to the previous papers, also performing experiments on the recently released XL-SUM dataset \cite{hasan-etal-2021-xl}.

\section{Analyzing Factual Consistency Models}
In this section, we present the factual consistency models we employ for detecting factual errors in the reference summaries. We introduce three state-of-the-art models that each rely on a distinctive mechanism. We rely on human annotations from the FRANK benchmark \cite{pagnoni-etal-2021-understanding} to validate the effectiveness of these models and analyze the variations in their performance when shifting dataset domain or error category. The FRANK benchmark is the most recent benchmark proposed for evaluating factual consistency models, it is the only benchmark to date that provides a categorization of factual errors.

\subsection{Factual Consistency Models}
\label{Factual Consistency Models}

In order to be comprehensive in our choice of factual consistency models, we run experiments on the top performing model from the FRANK benchmark as well as two other recently proposed models not included in the FRANK benchmark. Each of the three models we choose depends on a different mechanism, thus they are expected to complement each other in detecting different types of errors.

\paragraph{BERTScore\_Art}
\citet{zhang2019bertscore} introduce BERTScore as an automatic evaluation metric for text generation. It computes a similarity score for each token in the candidate summary with each token in the reference summary leveraging BERT embeddings. It obtains the final scores by performing a greedy matching between tokens from both texts. Here we employ a slightly modified version BERTScore\_Art, which directly compares a summary to the source article instead of the reference summary, for the sake of modeling factuality. We use the precision score which matches each token in the summary to a token in the article, instead of the recall score which does the opposite. This is the factuality model that obtained the highest correlation with human annotations in the FRANK benchmark \cite{pagnoni-etal-2021-understanding}.

\paragraph{BARTScore}
\citet{NEURIPS2021_e4d2b6e6} formulate the evaluation of generated text as a text generation task from pre-trained language models. The basic idea is that a high-quality summary should be easily generated based on the source article. The factuality score of a summary is calculated as its generation probability conditioned on the source document. The idea is operationalized using the encoder-decoder based model BART and thus named BARTScore. It builds a connection between the pre-training objectives and evaluation of text generation models. BARTScore is shown to produce state-of-the-art results for factuality evaluation on the SummEval Dataset \cite{fabbri-etal-2021-summeval}.

\paragraph{DAE (Dependency Arc Entailment)} \citet{goyal-durrett-2020-evaluating} propose to decompose summaries into dependency arcs and train an entailment model that makes independent factuality judgments for each dependency arc of the summary. The judgements made by the model are binary and we use the probability for the factual class in order to obtain a continuous score. The arc-level judgements are then aggregated into summary-level by computing the mean score for all arcs present in the summary. The initial work proposes to train the entailment model on synthetic datasets. However, they later extend their method by training it on human annotations and achieve improved performance \cite{goyal-durrett-2021-annotating}. This extended version of DAE is proven to be the best performing entailment-based model tested on the factuality dataset introduced in \citet{falke-etal-2019-ranking}.

\begin{figure}[t]
    \centering
    \includegraphics[scale=0.26]{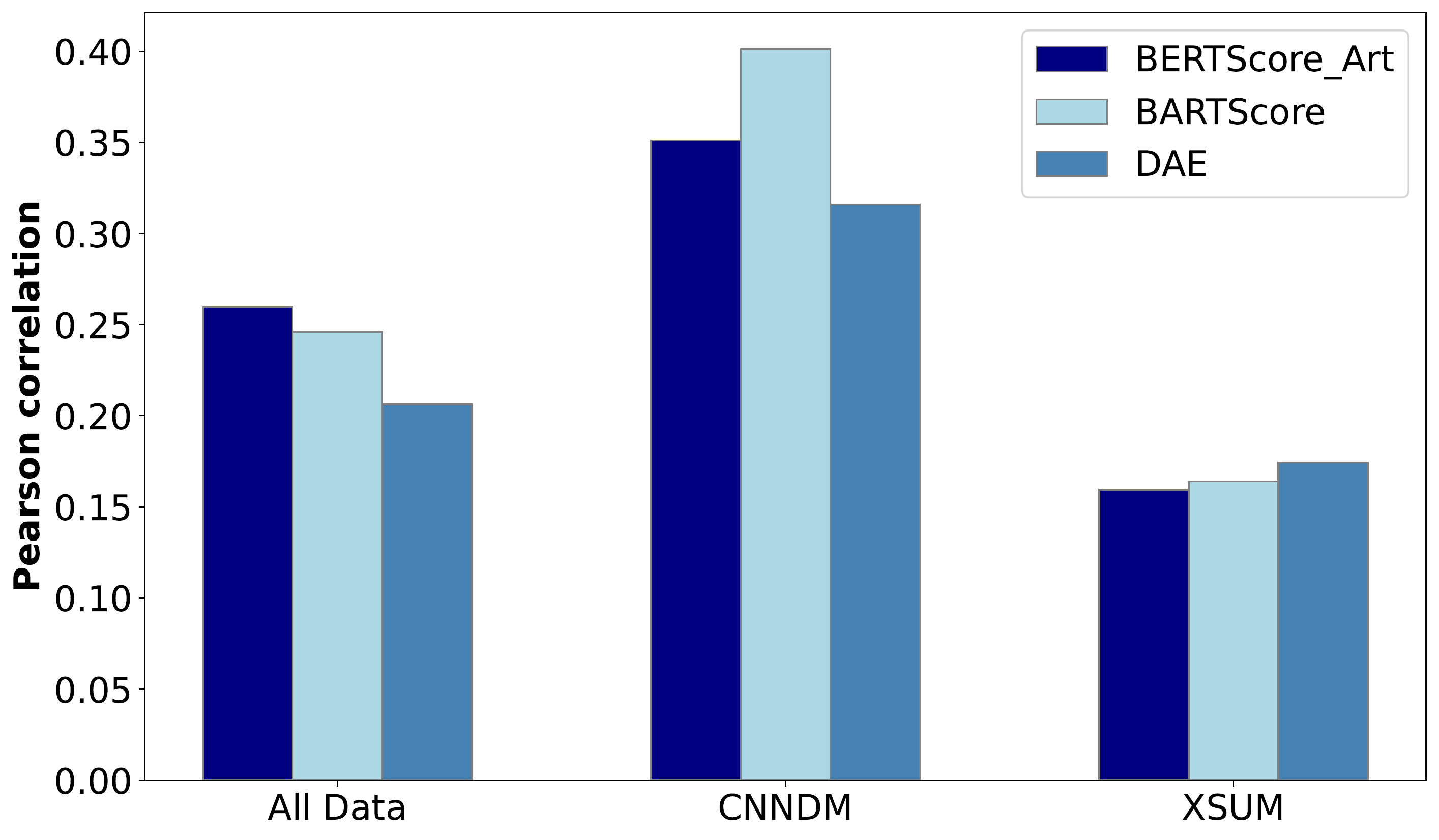}
    \caption{Partial Pearson correlation on different datasets. The three models demonstrate similar performance. All models have higher correlation on the CNN/DM dataset.}
    \label{fig:frank}
\end{figure}

\subsection{Model Validation Using the FRANK Benchmark}
\label{Analysis Using the FRANK Benchmark}

The FRANK benchmark proposes a linguistically motivated typology of factual errors for fine-grained analysis of factuality in summarization systems: semantic frame errors, discourse errors and content verifiability errors. Semantic frames refer to the schematic representation of an event, relation, or state and a \textit{semantic frame error} is an error that only involves participants of a single frame. \textit{Discourse errors} extend beyond single semantic frames and consist of erroneous relations between different discourse segments. \textit{Content verifiability errors} arise when information in the summary cannot be verified against the source document. 

The FRANK benchmark consists of human-annotated categorical error scores for 2250 model summaries. We use these annotations to compute partial correlation scores with human judgements for different models on different datasets and error types. We now examine these correlations.

\begin{figure*}[!ht]
	\centering
	\subfloat[XSUM]{
	\includegraphics[scale=0.26]{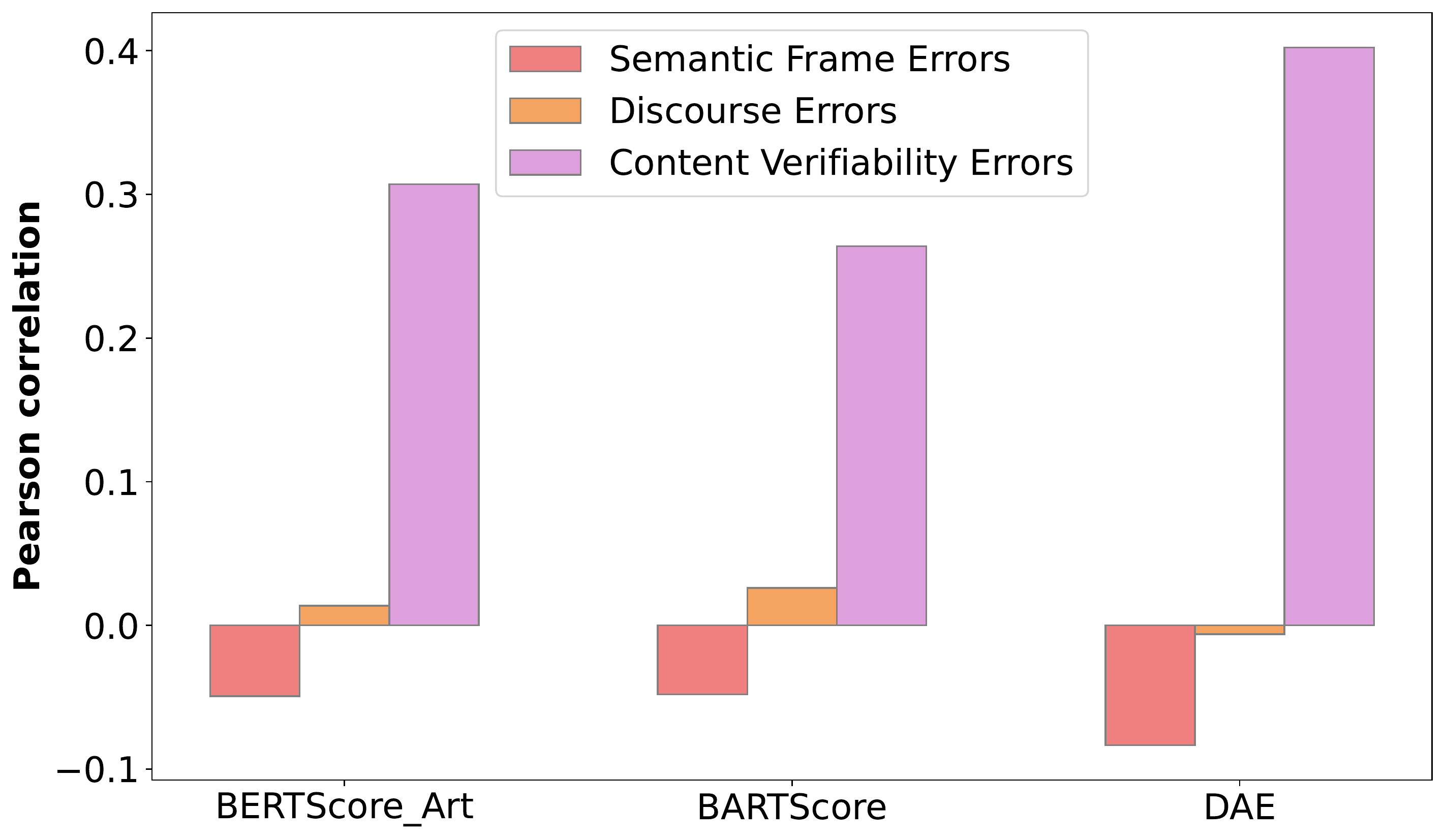}
	}
	\subfloat[CNN/DM]{
	\includegraphics[scale=0.26]{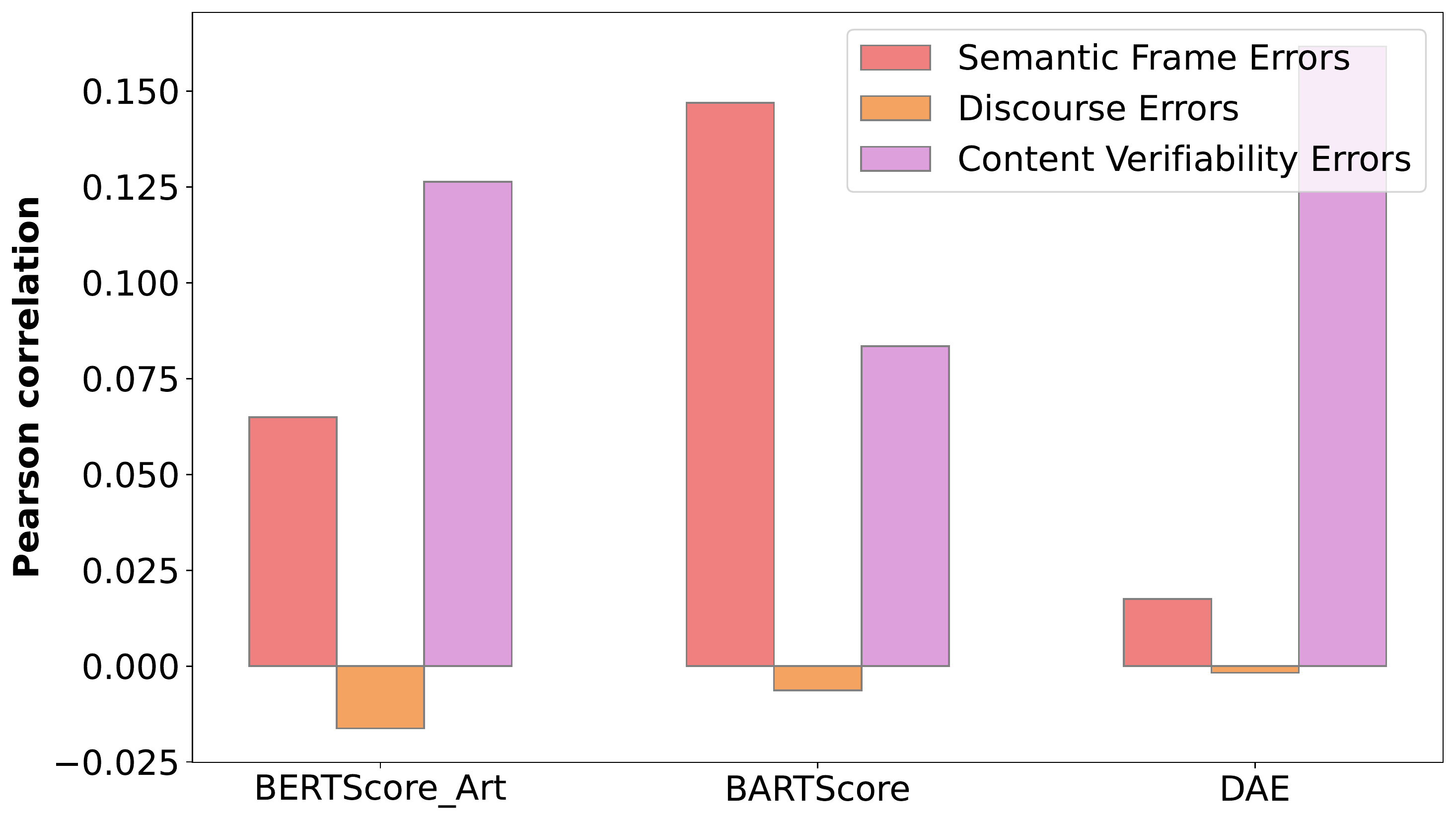}
	}
	\caption{Negative difference in partial Pearson correlation when flipping labels of an error type. Higher value indicates higher influence of the given error type in the overall correlation.}
	\label{fig:frank_cat}
\end{figure*}

\paragraph{Partial Pearson correlation on different datasets.}
Figure \ref{fig:frank} shows the partial pearson correlation between different models and human judgments on document-summary pairs from different datasets. The FRANK benchmark includes 1250 pairs from the CNN/DM dataset and 1000 from the XSUM dataset. We observe that the three models achieve comparable performance across all datasets with BARTScore obtaining the best performance on CNN/DM, DAE obtaining the best performance on XSUM and BERTScore\_Art on both of them combined.

\paragraph{Partial Pearson correlation on different error types.}
Figure \ref{fig:frank_cat} shows the variation in partial Pearson correlation when flipping the labels of a specific error type. A higher positive bar indicates that the given type of error has high influence in the overall correlation. In other words, the factual consistency model performs well in detecting an error type if it has a high positive bar. We observe that DAE performs the best for content verifiability errors on both of the datasets, but gives unsatisfactory results for the other two error types. This is expected as DAE scores each dependancy arc independantly and thus cannot detect erros spanning across different arcs. BERTScore\_Art and BartScore are both able to achieve positive results for discourse errors on XSUM, and for semantic frame errors on CNN/DM. However, it is worth pointing out that all of the three models generally perform the worst for discourse errors, indicating that this is a challenging future direction for work on factuality models. 

Due to the relative strength of each model in identifying different types of errors on different datasets, we choose to use all three of them in analyzing the summarization datasets in Section \ref{Examining Summarization Benchmarks}. We further combine these three models in our filtration process in Section \ref{Filtration}.

\section{Examining Summarization Benchmark Datasets}
\label{Examining Summarization Benchmarks}
To question the validity of current summarization benchmark datasets, we first provide an overview of the context in which each dataset was created and the methodology employed in the construction procedures. In addition, we make use of the previously mentioned factual consistency models to evaluate the factuality of human reference summaries collected in each dataset. Our selection of summarization datasets is based on their popularity in recently published papers introducing new summarization systems. We believe that it is crucial to examine their validity as they play a fundamental role in both the development and evaluation of new systems. The most popular benchmark datasets according to our criteria are CNN/DM \cite{NIPS2015_afdec700} and XSUM \cite{narayan-etal-2018-dont}. We also include the very recently released dataset XLSumm \cite{hasan-etal-2021-xl}. It is not yet widely employed but including it alongside the other two can help us inspect the most recent advances made in the field of summarization dataset creation. \textit{Does the most recent benchmark dataset exhibit any improvement on factuality in comparison to the previous ones with well-known issues?}

\subsection{Summarization Benchmarks}
\label{Summarization Benchmarks}
We present the three summarization benchmark datasets in the order of their time of release. We observe an evolution in the construction methodology over time.

\begin{figure*}[!ht]
	\centering
	\subfloat[BERTScore]{
	\includegraphics[scale=0.32]{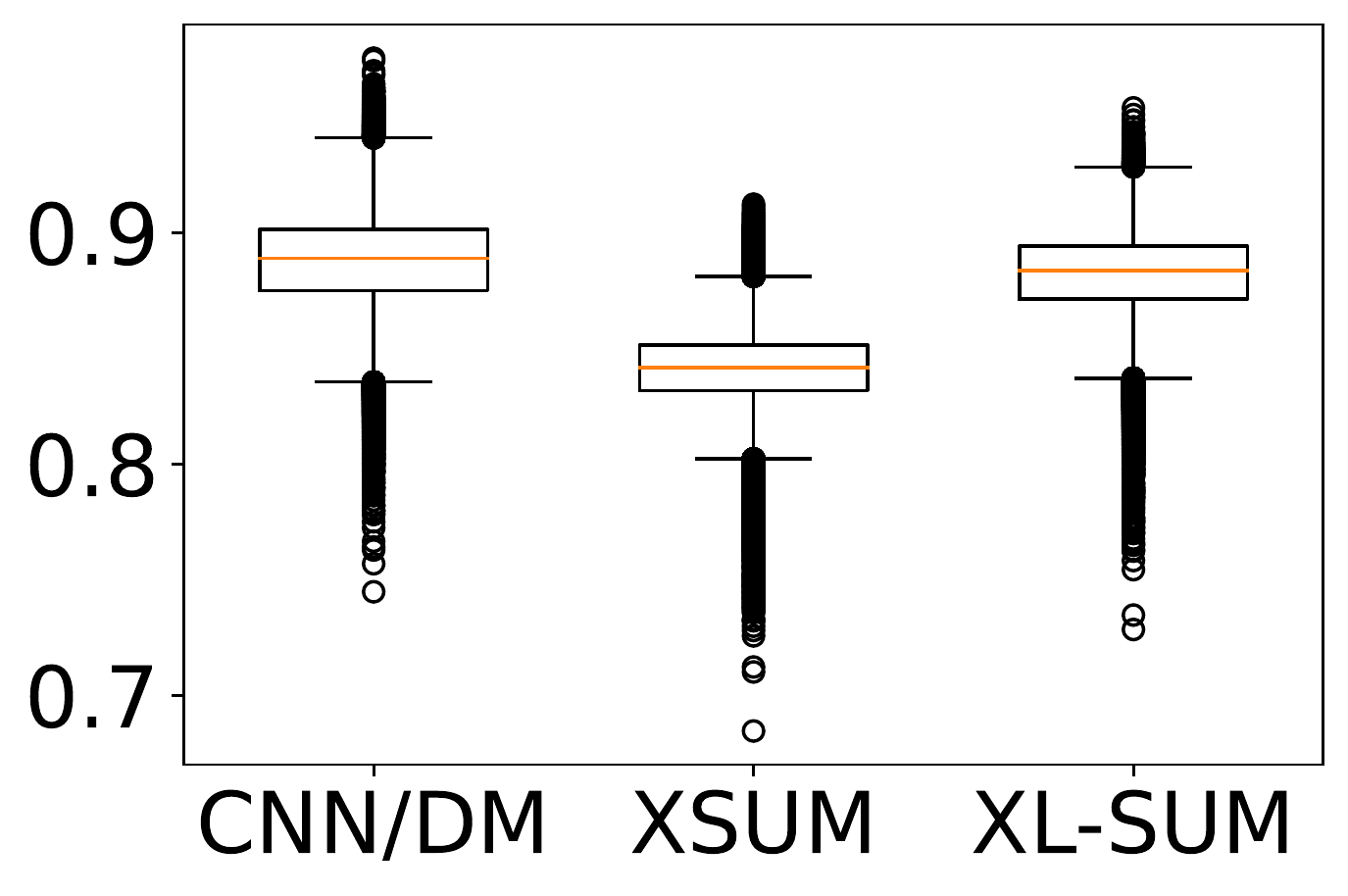}
	}
	\subfloat[BARTScore]{
	\includegraphics[scale=0.32]{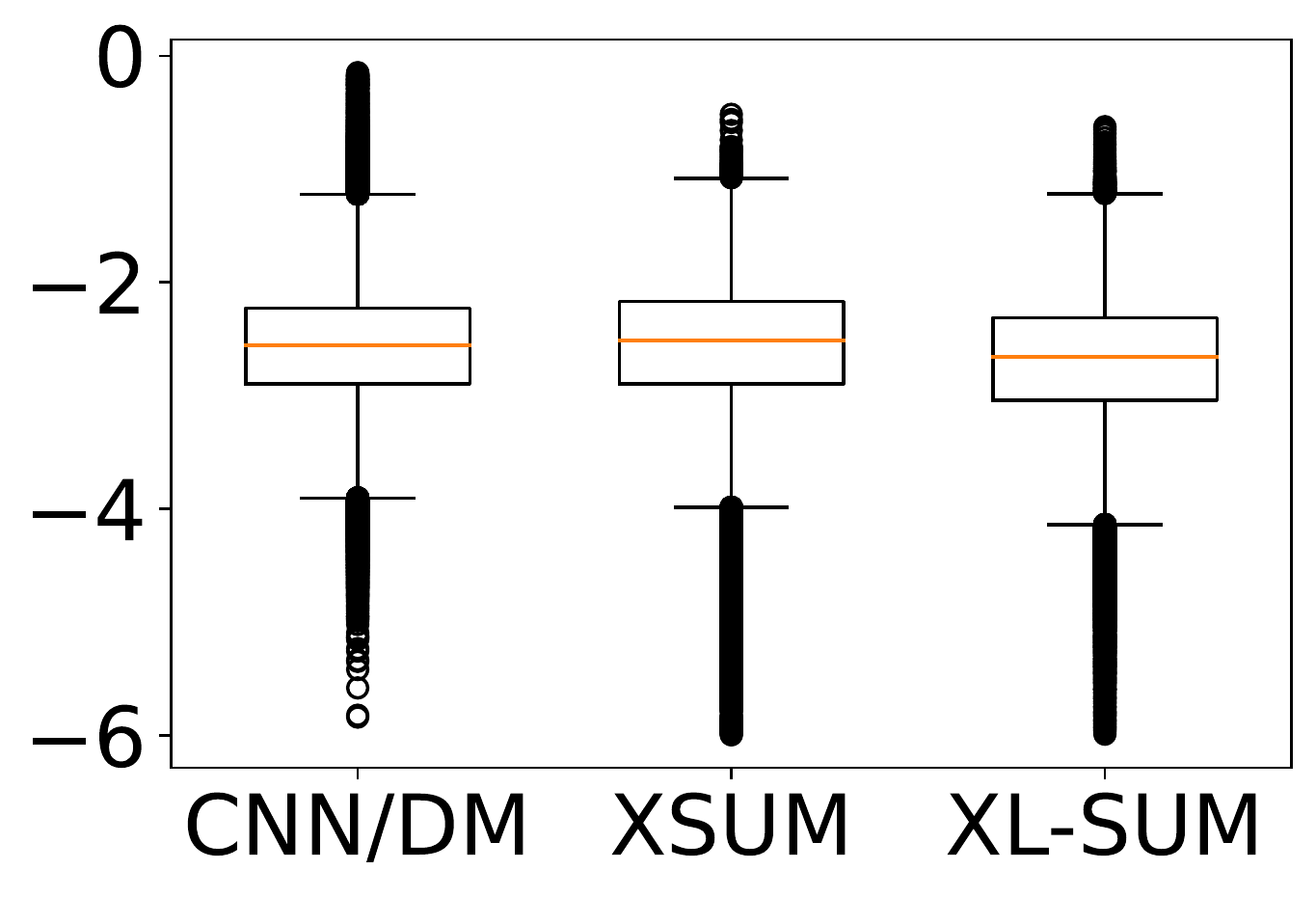}
	}
	\subfloat[DAE]{
	\includegraphics[scale=0.32]{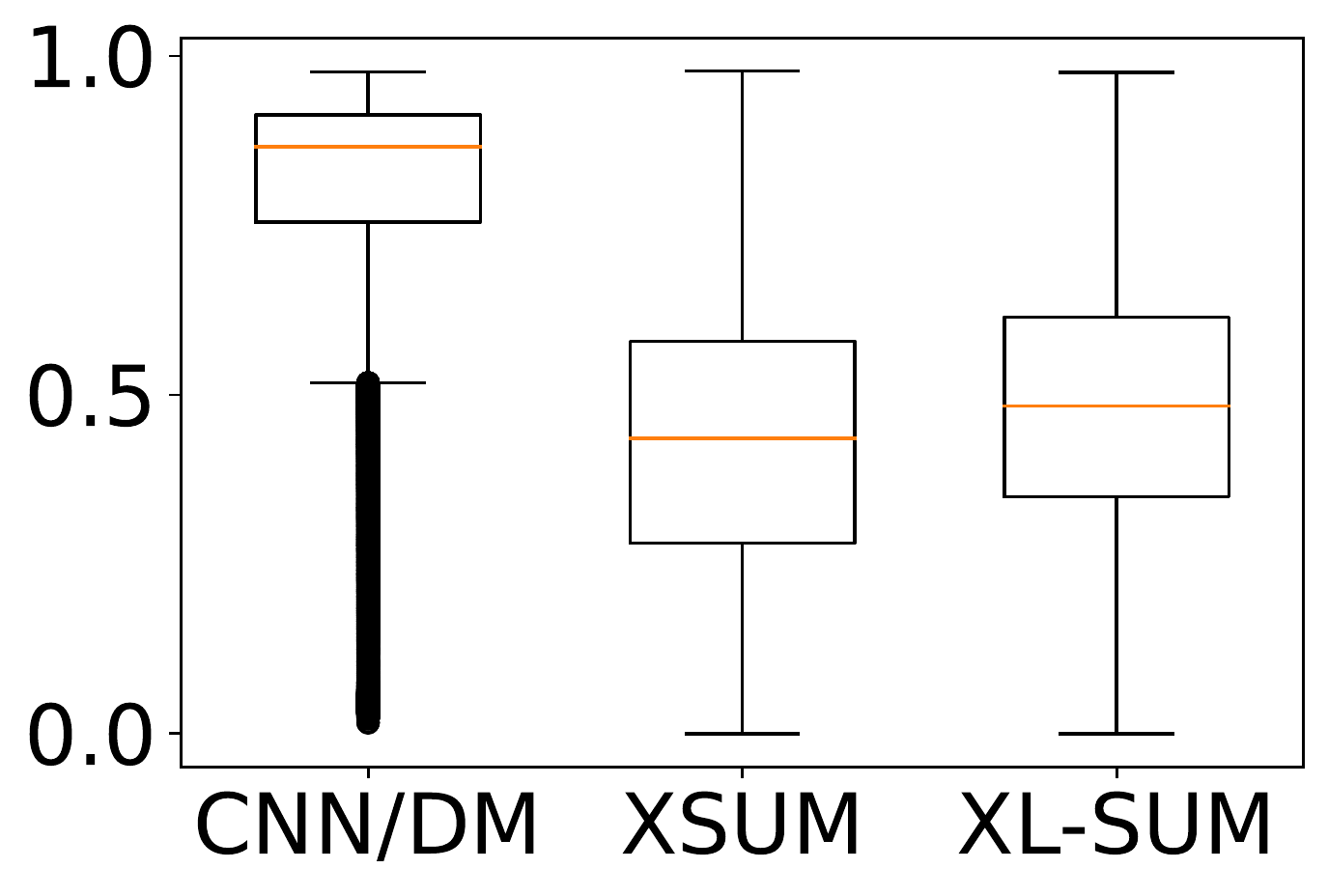}
	}
	\caption{Box plot of factuality scores obtained for the three datasets.}
	\label{boxplot}
\end{figure*}

\paragraph{CNN/DM}
The CNN/DM dataset \cite{NIPS2015_afdec700} was initially constructed as a Question Answering dataset composed of newswire articles in English and their corresponding highlights from the two platforms CNN and Daily Mail. \citet{cheng-lapata-2016-neural} later converted it into a summarization dataset by simply concatenating these highlights into summaries. It has now become the most broadly employed summarization dataset for the English language. However, the summaries formed from the concatenation of bullet points exhibit low degrees of abstraction and coherence, which are both highly desirable qualities for abstractive summarization systems. The dataset consists of 311,971 document-summary pairs. 

\paragraph{XSUM}
\citet{narayan-etal-2018-dont} create another large-scale dataset for abstractive summarization by crawling online articles from the BBC platform. They take the first line of an article as the summary and the rest of the article as the source document. This method for annotating summaries guarantees high levels of abstraction but creates other issues as the first line of an article is often not written to be the summary. It might either include meta-information such as the author and publication date or serve as background introduction thus containing information never mentioned again in the rest of the article. XSUM contains 226,711 document-summary pairs.

\paragraph{XL-Sum}
 XL-Sum \cite{hasan-etal-2021-xl} is a recently introduced abstractive summarization dataset containing document-summary pairs also extracted from the BBC platform. The automatic annotation strategy of summaries is similar to that of XSUM. However, they find the first line of many articles to contain meta-information and thus annotate the bold paragraphs instead as the summaries. The dataset covers 44 languages, for many of which no public summarization datasets were previously available. Summaries contained in XL-Sum are also highly abstractive. The English subset of this dataset contains 329,592 document-summary pairs.
 
\subsection{Factual Consistency of Summarization Benchmarks}

\label{Factual Consistency of Summarization Benchmarks}

\begin{table*}
\centering
\resizebox{\textwidth}{!}{\begin{tabular}{lr|ccccc}
\hline\hline
& &\#Samples & Selection Ratio & Document Length & Summary Length & \\
\hline
\textbf{CNN/DM}  & Full train set & 287,113  & * & 788 & 54 & \\
 & SummFC selection & 153,666 & 53.52\% & 726 & 56 & \\
\hline
\textbf{XSUM}  & Full train set & 204,045 & * & 430 & 23 &  \\
 & SummFC selection & 118,427 & 58.03\% & 405 & 23 &  \\
\hline
\textbf{XLSUM} & Full train set & 306,522 & * & 530 & 24 &  \\
 & SummFC selection & 161,200 & 52.59\% & 491 & 24 &  \\
\hline\hline
\end{tabular}}
\caption{\label{table:statistics}
Statistics of the SummFC dataset.
}
\end{table*}

We focus on evaluating the factual consistency of human reference summaries contained in each of the summarization benchmark datasets. We often assume that human references are ``gold standards'' but there is a lot to question in their validity considering how they are annotated. 

Here, we perform an analysis on the human references using the three factuality models introduced in Section \ref{Factual Consistency Models}: BERTScore\_Art, BARTScore and DAE. The results are shown in Figure \ref{boxplot}. We remark that factuality scores produced by all three models rank CNN/DM as the most factual dataset by a large margin. This is due to its low level of abstraction and thus do not indicate its superiority as an abstractive summarization dataset. However, it is also worth pointing out that the improvement achieved betweeen XSUM and the recently created XL-SUM is not significant. By qualitatively analyzing the lowest ranking summaries scored by each model from each dataset, one can confirm their non factuality. A representative non factual example for XSUM is shown in Table \ref{table:examples}. Examples for the two other datasets are shown in Appendix Section \ref{sec:appendix}.

\begin{figure*}[!ht]
	\centering
	\includegraphics[scale=0.48]{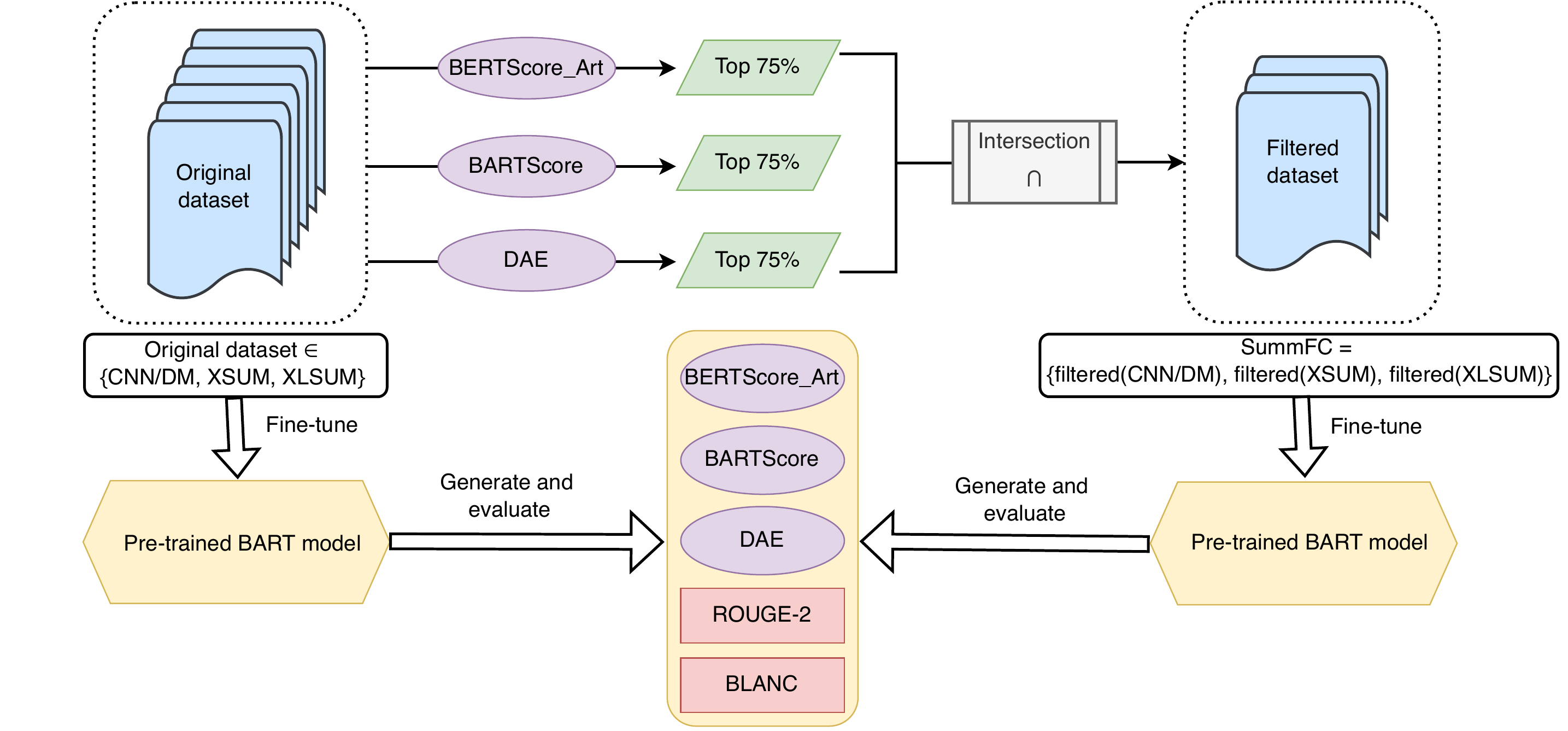}
	\caption{Pipeline for creating and evaluating the SummFC dataset.}
	\label{fig:pipeline}
\end{figure*}

\begin{table*}
\centering
\resizebox{\textwidth}{!}{\begin{tabular}{lr|cccccc}
\hline\hline
& & \textbf{BERTScore\_Art} & \textbf{BARTScore} & \textbf{DAE} & \textbf{BLANC} & \textbf{ROUGE-2} & \\
\hline
\textbf{CNN/DM}  & Random 53.52\%  & 0.9332 & -2.662 & 0.9046 & 0.1538 & 22.80 & \\
  & Full train set & 0.9346 & \textbf{-2.577} & 0.9001 & \textbf{0.1593} & \textbf{23.77} & \\
 & SummFC selection & \textbf{0.9364} & -2.600 & \textbf{0.9073} & 0.1556 & 23.67 & \\
\hline
\textbf{XSUM}  & Random 58.03\%  & 0.8973 & -2.386 & 0.4134 & 0.07288 & 20.27 & \\
 & Full train set & 0.8944 & -2.452 & 0.4201 & 0.07316 & \textbf{21.78} & \\
 & SummFC selection & \textbf{0.8982} & \textbf{-2.383} & \textbf{0.4473} & \textbf{0.07378} & 21.50 & \\
 
\hline
\textbf{XLSUM}  & Random 52.59\%  & 0.8979 & -2.537 & 0.4612 & 0.06872  & 19.63 \\
  & Full train set & 0.8961 & -2.594 & 0.4695 & 0.06886 & \textbf{20.99} & \\
 & SummFC selection & \textbf{0.8989} & \textbf{-2.501} & \textbf{0.5120} & \textbf{0.06983} & 20.78 & \\
\hline\hline
\end{tabular}}
\caption{\label{table:results}
Results for BART models fine-tuned on different selections of datasets. The statistical significance ($p < 0.05$) of all results are confirmed using the Wilcoxon signed-rank test.
}
\end{table*}

\begin{figure*}[h]
	\centering
	\subfloat[BLANC]{
	\includegraphics[scale=0.25]{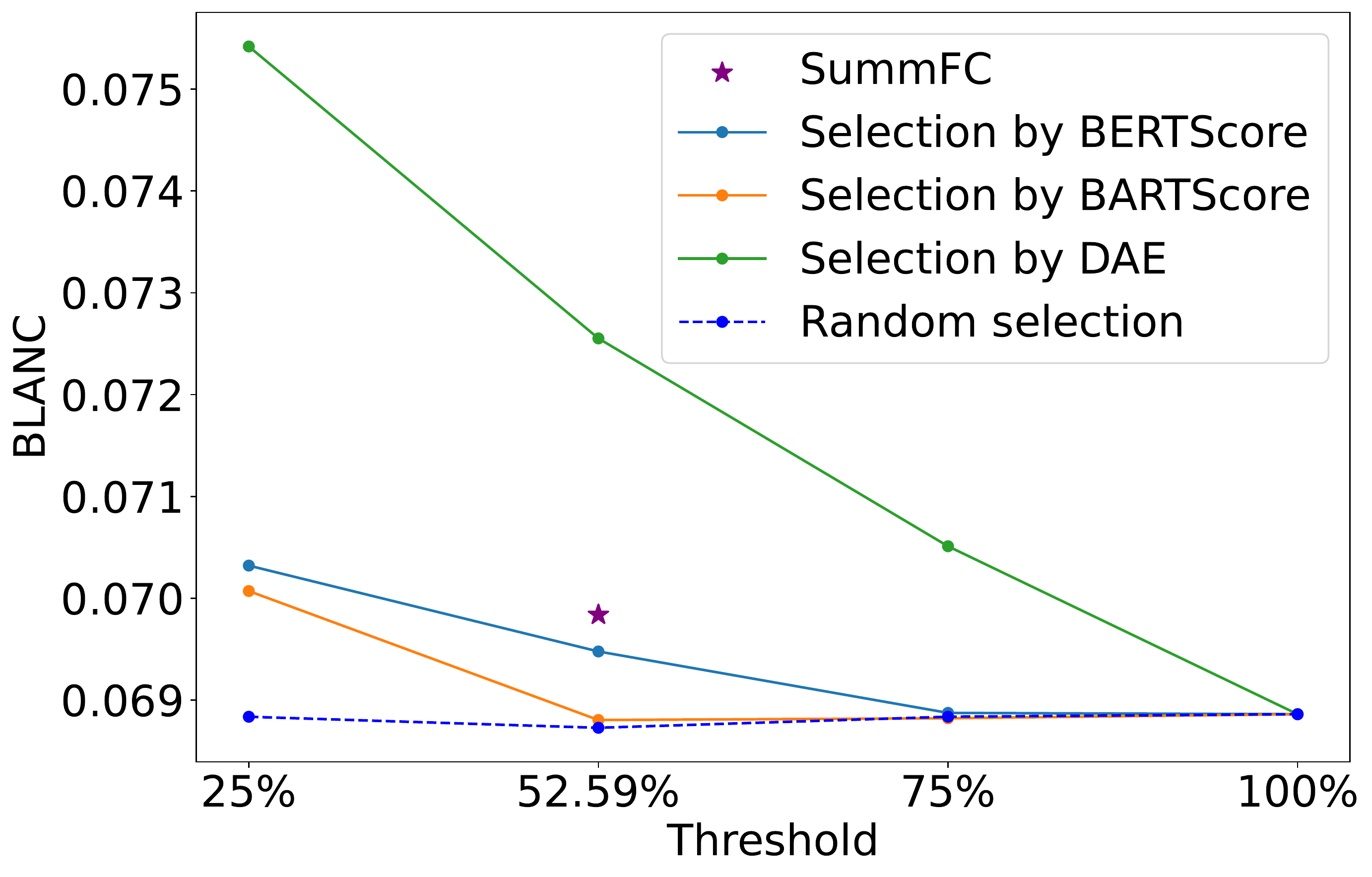}
	}
	\subfloat[ROUGE-2]{
	\includegraphics[scale=0.25]{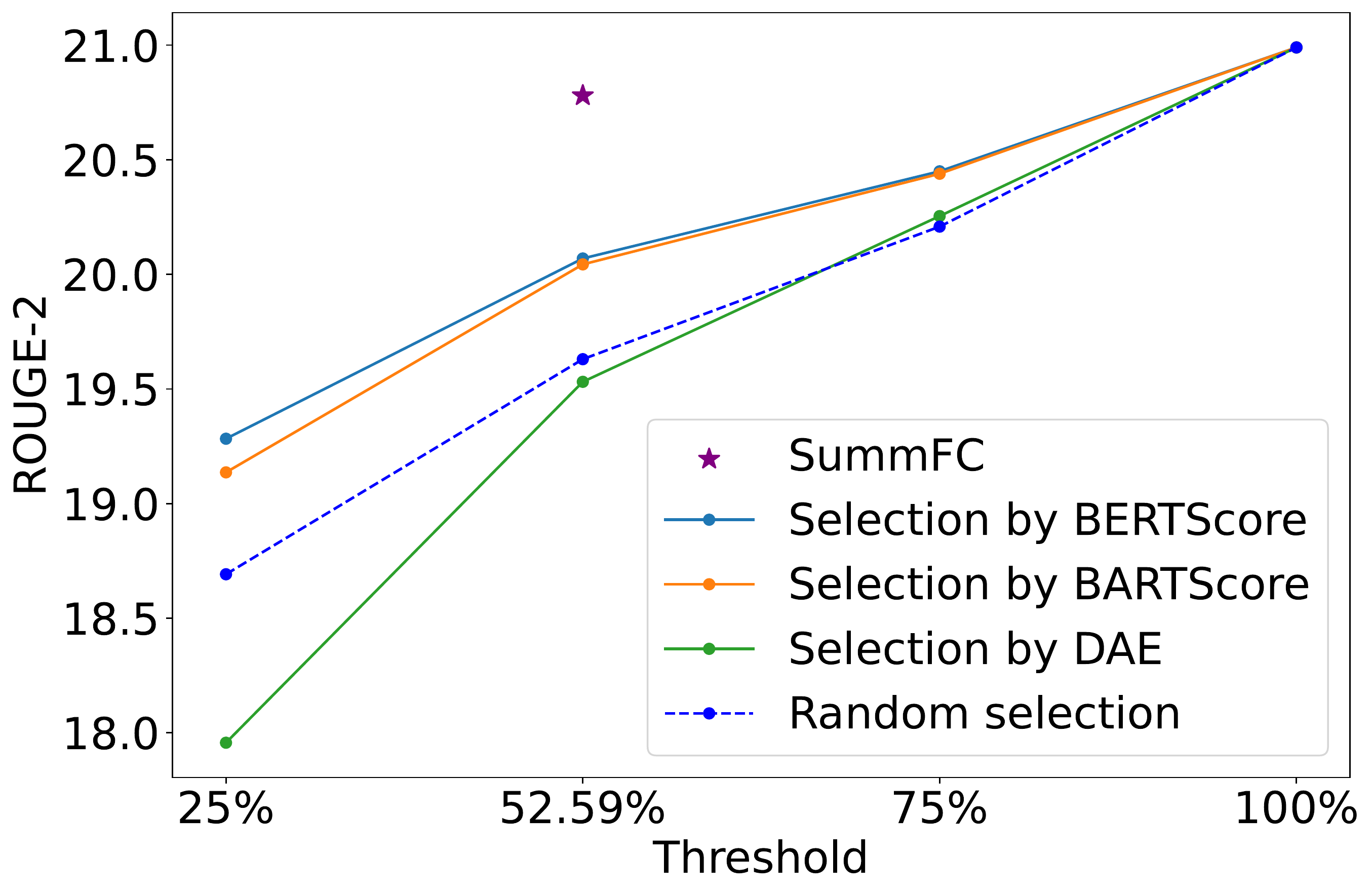}
	}
	\caption{Results for BART models fine-tuned on different subsets of the XLSUM dataset, filtered by different factuality models with varying thresholds.}
	\label{thresholds}
\end{figure*}

\section{Introducing the SummFC Dataset}

The approach we choose to address the factuality problem in these summarization datasets is to filter the samples by their factual consistency scores obtained in Section \ref{Factual Consistency of Summarization Benchmarks}. We apply this idea to construct \textbf{SummFC}, the new \textbf{Summ}arization benchmark dataset with improved \textbf{F}actual \textbf{C}onsistency. In the remainder of this section, we first present our filtration methodology, followed by statistics of the SummFC dataset. We then introduce the evaluation metrics used in the comparison of summaries generated by models trained on the original benchmarks and models trained on SummFC. Finally, we perform experiments on the three datasets introduced in Section \ref{Summarization Benchmarks} and observe advantages of the SummFC dataset. 

\subsection{Filtration Methodology}
\label{Filtration}
In Figure \ref{fig:pipeline}, we illustrate our pipeline for filtration and evaluation of the three datasets. The filtration methodology is based on the factuality models BERTScore\_Art, BARTScore and DAE. For each of the three datasets, we filter out the bottom 25\% of document-summary pairs scored by each of the three models. In other words, we only keep the intersection between the top 75\% of samples scored by each factuality model. This choice is made because the three factuality models are shown to complement each other in the detection of erroneous samples (see Section \ref{Analysis Using the FRANK Benchmark}). Experiments in Section \ref{Experiments and Results} show that datasets filtered by combined models outperform single model filtration with the same threshold.

\subsection{The SummFC Dataset}
In Table \ref{table:statistics}, we show statistics for the SummFC datasets. For all three datasets, the selection ratio of samples to retain is between 50\% and 60\%. In the context of our filtration methodology removing the lowest scored 25\% of samples for each factuality model, this final selection ratio proves that there is a high degree of overlap between the samples different factuality models choose to filter out. We also report the average length of documents and summaries in each dataset. We compute length as the number of words (instead of tokens) included in the text. We might expect to achieve higher factual consistency scores for shorter summaries, as they naturally contain less information. However, we show that our filtration methodology does not particularly favor samples with shorter reference summaries.

\subsection{Evaluation Metrics}
\label{Evaluation Metrics}

To compare the quality of summaries produced by models fine-tuned on the the original datasets and the SummFC selection, we employ the following evaluation metrics capturing different quality aspects.

\paragraph{Factual Consistency Models}
For testing the improvement in \textbf{factual consistency} of the produced summaries, we employ the same factuality models as we use for filtration in Section \ref{Factual Consistency Models}: BERTScore\_Art, BARTScore and DAE.

\paragraph{BLANC}
BLANC \cite{vasilyev-etal-2020-fill} is a reference-free evaluation metric for summarization. It is defined as a measure of how well a summary helps an independent, pre-trained language model while it performs its language understanding task (masked token task) on a document. In other words, BLANC measures the \textbf{informativeness} of a summary in the context of document understanding. Here, we use the BLANC-help version, which uses the summary by concatenating it to each document sentence during inference.

\paragraph{ROUGE-2}
ROUGE-2 \cite{lin-2004-rouge} is a reference-based text generation metric computing the bigram overlap between reference and candidate summaries. While reference-free evaluation metrics are advantageous due to the issues in reference summaries, there are still quality aspects it cannot cover. There is currently no reference-free metric that is able to measure a summary's \textbf{saliency} (\ie~coverage of salient information). Thus we still need to compare the generated summaries to the reference ones with ROUGE-2, assuming that the reference summaries exhibit a high degree of saliency. We also process the samples in the test set with our filtration methodology when using ROUGE-2. For all the other evaluation metrics, we do not perform filtration for the test set because they are reference-free and thus not influenced by misleading reference summaries.

\subsection{Experiments and Results}
\label{Experiments and Results}

We fine-tune the pre-trained BART-base model \cite{lewis-etal-2020-bart} from the Transformers library \cite{wolf-etal-2020-transformers} on the different summarization datasets. Since our goal is not to advance summarization systems but to compare different benchmark datasets, we do not experiment with hyperparameter tuning. For all parameters except batch size, we use the default settings in the Transformers library. Both the training and prediction are done on a NVIDIA TITAN V GPU with a batch size of 8. We present results for models fine-tuned on the original benchmarks and SummFC in Table \ref{table:results}. We also create a random selection baseline for each dataset by uniformly sampling a random subset of samples with equal size to the SummFC selection. 

We observe that models trained on the SummFC selection of XSUM and XLSUM achieve the highest scores for all reference-free evaluation metrics. Although the selection procedure is only based on factual consistency, we see that the summarization systems have also improved in the informativeness aspect as measured by BLANC. For the reference-based metric ROUGE-2, models trained on SummFC obtain comparable scores to those trained on the original dataset and beat the random baseline by a large margin. It is also interesting to remark that for some of the factuality scores, even the random baseline achieves better performance than the full train set. This further confirms the conclusion that too much erroneous training data hinders factual consistency in summarization models. The results on CNN/DM are consistent with the other two datasets, except for BARTScore and BLANC. We believe that our filtration methodology is slightly less effective on CNN/DM due to the fact that this dataset manifests the lowest degree of factual inconsistency. Another important reason is that the BART model used in BARTScore is also fine-tuned on CNN/DM, which makes it biased on this dataset. However, it is worth noting that SummFC is considerably smaller in size compared to the original ones. This means that \textit{using SummFC, we can achieve better results on nearly all quality aspects while reducing training time and lowering the need for computational resources}. 

Figure \ref{thresholds} shows metric scores obtained when fine-tuning on filtered XLSUM datasets with single factuality models at different thresholds. We show scores for BLANC and ROUGE-2 because these two metrics were not used during filtration. We also create a baseline by randomly selecting samples with proportions equal to the tested thresholds. As the filtration criteria becomes stricter, BLANC scores increase while ROUGE-2 scores decrease. Our final filtration threshold is chosen as a compromise between the optimization of these two scores. We also observe that there is no single factuality model that performs the best for both scores. \textit{In general, the SummFC combined filtration strategy outperforms single factuality models at the same threshold.}

\section{Conclusion}
In this paper, we demonstrate that popular summarization datasets suffer from the lack of factual consistency and that summarization models trained on these datasets are not adequate for the task of abstractive summarization. We show that this problem can be solved by filtering the benchmark datasets with scores from factual consistency models. We propose a filtration methodology combining three state-of-the-art factual consistency models and introduce the SummFC dataset. SummFC is a unified \textbf{Summ}arization benchmark consisting of \textbf{F}actually \textbf{C}onsistent samples chosen from CNN/DM, XSUM and XLSUM. Experiments indicate that, in general, models trained on the smaller SummFC generate summaries with higher quality than models trained on the larger original datasets. Our findings suggest that more deliberate considerations should be made in the construction of benchmark datasets and that continuous revisions for the already existing ones are particularly necessary. However, constructing adequate benchmark datasets with textual contents and labels matching the initial formulation of NLP tasks remains an open question, major obstacle, and unresolved issue for the whole community.

\section*{Limitations}
In this work, we restrict ourselves to the most popular summarization datasets. The three analyzed datasets share many similarities with each other and thus cannot account for diversity. They are all single-document single-reference English language summarization datasets in the news domain. Due to the limited sequence length accepted by Transformer-based factual consistency models, our filtration methodology cannot be generalized for longer document-summary pairs which are frequently present in other domains such as scientific articles and creative writing. The methodology also cannot be generalized for other languages. Factual consistency models such as DAE are trained on annotated datasets which only exist for the English language. This provides motivation for improving the generalization of factual consistency models.

Another limitation of our work is the lack of human evaluation. We prove the superiority of the SummFC dataset by evaluating the generated summaries with automatic metrics. Although these metrics achieve state-of-the-art correlations with human annotated scores, they are still known to significantly differ from human judgements. While the factual consistency models that we employ also represent the current state-of-the art, it is far from guaranteed that they are able to identify all the factuality errors. It is only with human evaluation that we can provide a complete picture of where SummFC falls on the full dataset quality continuum. There is undoubtedly still more work to be done in continuing to refine datasets to actually measure summarization.

\section*{Ethics Statement}
We believe that research on improving the factual consistency of summarization systems can create positive social impact. We live in an era of information explosion and everyone, to some degree, relies on summarization to process the information overload. It is our responsibility to guarantee the greater public's access to truthful information.

Our research also brings positive impact on the environmental aspect. Training on smaller and higher quality datasets significantly reduces the consumption of computational energy while also boosting performance.

\section*{Acknowledgements}
We thank the anonymous reviewers for their helpful feedback and insightful comments. The first and last authors were supported by the ANR HELAS chair (ANR-19-CHIA-0020-01). The second author was supported by the Télécom Paris research chair on Data Science and Artificial Intelligence for Digitalized Industry and Services (DSAIDIS).

\bibliography{anthology,custom}
\bibliographystyle{acl_natbib}

\appendix

\vspace{1cm}

\section{Examples of Factually Inconsistent Reference Summaries}
\label{sec:appendix}

\begin{table}[h]
\begin{center}
\begin{tabular}{ |p{0.95\linewidth}| } 
 \hline
 \textsl{\textcolor{blue}{Source Document: }} \\
 \textsl{By . Leah Simpson . PUBLISHED: . 16:46 EST, 19 July 2012 . | . UPDATED: . 02:31 EST, 20 July 2012 . With the season finale airing on Sunday night, The Bachelorette star Emily Maynard is already making arrangements to extend her 15 minutes of fame - with a move to Hollywood on the cards.}
  \textsl{......}
 \textsl{But she's really excited to get the date, location and all of the details set so that she can marry Jef and having it air on TV fits in perfectly with her plans.' Happy couple: Emily is engaged to get married to Jef Holm apparently . }  \\
 \hline
 \textsl{\textcolor{blue}{Factually Inconsistent Reference Summary: }} \\
 \textsl{\textcolor{red}{Bachelorette spoiler alert .}}\\
 \hline
\end{tabular}
 \caption{\label{table:examples1}
Example of a reference summary from the CNN/DM dataset. It is more of a click-bait title for attracting attention than a real summary.}
\end{center}
\end{table}

\begin{table}[h]
\begin{center}
\begin{tabular}{ |p{0.95\linewidth}| }
 \hline
 \textsl{\textcolor{blue}{Source Document: }} \\
 \textsl{Bosses said the move was part of an efficiency drive, with 10 posts set to go in Haverfordwest and 20 at its plant in Aspatria, Cumbria. "We recognise that the impact of these proposed changes is significant for the people affected and we are committed to treating people with respect and consideration," said a spokesman. A staff consultation starts this week.}\\
 \hline
 \textsl{\textcolor{blue}{Factually Inconsistent Reference Summary: }} \\
 \textsl{\textcolor{red}{A total of 30 jobs are under threat at two First Milk creameries in Pembrokeshire and the Lake District.}}\\
 \hline
\end{tabular}
 \caption{\label{table:examples2}
Example of a reference summary from the XL-SUM dataset. Information in summary not explained in the source document. }
\end{center}
\end{table}

\end{document}